\documentclass{article}

\usepackage[nonatbib, preprint]{neurips_data_2024}

\usepackage[utf8]{inputenc} 
\usepackage[T1]{fontenc}    
\usepackage{hyperref}       
\usepackage{url}            
\usepackage{nicefrac}       
\usepackage{graphicx}
\usepackage{color}
\usepackage{enumitem}

\usepackage{amsmath}
\usepackage[capitalize]{cleveref}
\usepackage{nicefrac}

\usepackage{pdflscape}
\usepackage{afterpage}

\newcommand{\emocam}{EmoCAM\mbox{++}}

\newcommand{\figref}[1]{Figure~\ref{#1}}

\newcommand{\fmriimgs}{the fMRI stimuli}
\newcommand{\supscr}[1]{$^{\text{#1}}$}

\title{Assessing the Alignment of Popular CNNs to the Brain for Valence Appraisal}

\author{%
  Laurent Mertens$^{1,2}$  \\
  \And Elahe' Yargholi$^3$ \\
  \And Laura Van Hove$^{3,4}$ \\
  \And Hans Op de Beeck$^3$ \\
  \AND Jan Van den Stock$^4$ \\
  \And Joost Vennekens$^{1,5}$ \\
  \AND \\
  $^1$KU Leuven, De Nayer Campus, Dept. of Computer Science \\
  J.-P. De Nayerlaan 5, 2860 Sint-Katelijne-Waver, Belgium \\
  $^2$ Leuven.AI - KU Leuven Institute for AI, 3000 Leuven, Belgium \\
  $^3$Department of Brain and Cognition, Leuven Brain Institute,\\Faculty of Psychology~\&~Educational Sciences\\
  KU Leuven, 3000 Leuven, Belgium \\
  $^4$Neuropsychiatry, Leuven Brain Institute \\
  KU Leuven, 3000 Leuven, Belgium \\
  $^5$Vrije Universiteit Brussel, Brussels, Belgium \\
  \texttt{laurent.mertens@kuleuven.be} \\
}

\begin{document}

\maketitle

\begin{abstract}
Convolutional Neural Networks (CNNs) are a popular type of computer model that have proven their worth in many computer vision tasks. Moreover, they form an interesting study object for the field of psychology, with shown correspondences between the workings of CNNs and the human brain. However, these correspondences have so far mostly been studied in the context of general visual perception. In contrast, this paper explores to what extent this correspondence also holds for a more complex brain process, namely social cognition. To this end, we assess the alignment between popular CNN architectures and both human behavioral and fMRI data for image valence appraisal through a correlation analysis. We show that for this task CNNs struggle to go beyond simple visual processing, and do not seem to reflect higher-order brain processing. Furthermore, we present Object2Brain, a novel framework that combines GradCAM and object detection at the CNN-filter level with the aforementioned correlation analysis to study the influence of different object classes on the CNN-to-human correlations. Despite similar correlation trends, different CNN architectures are shown to display different object class sensitivities.
\end{abstract}

\section{Introduction}
Convolutional Neural Networks (CNNs) have been around for decades, and, inspired by the workings of the human visual system, are still a popular type of network for multiple computer vision tasks such as object detection and segmentation \cite{2024_CNNsForVision}. They also represent an interesting study object for the field of psychology, particularly for modeling the human visual system. Indeed, CNNs have been shown to exhibit similarities with visual processing in the human brain, in particular showing a hierarchical correspondence for object representation between lower/higher level visual cortex regions and lower/higher level CNN layers \cite{Yamins2014, Agrawal2014, Cadieu2014, Kubilius2016DeepNN, KubiliusSchrimpf2019CORnet, Nonaka2020BrainHS, CnnsForVisionNeuroscience}. More recent work however \cite{ReassesHierarchy}, using so-called ``direct interfacing'' whereby functional Magnetic Resonance Imaging (fMRI) data is directly fed as input to a CNN layer through a linear map, found that all visual ventral stream brain regions, including V1, best drove later CNN layers, concluding that all regions contained higher-level information. Others \cite{Xu2021LimitsTV, Mocz2023MultipleVO} highlight the limits to using CNNs as models of human vision. These investigations however all focus on fundamental visual perception, rather than on higher-level cognitive tasks.

In this paper, we investigate whether this correspondence to the brain also holds for more complex tasks such as social cognition, and in particular valence appraisal, i.e., how positive/pleasant or negative/unpleasant a certain scene or situation is. Recent work from psychology \cite{Yargholi2025} indicated that valence processing in the human brain of static images, all depicting people in various settings, is a complex and distributed affair. Distinguishing between people, scene and (whole) image valence, and dividing the brain into three major regions spanning the range from low-level visual regions to high-level anterior cortex regions, it was shown that higher level processing is needed to fully appraise an image's valence, with the low-level regions only sensitive to scene valence (i.e., contextual elements). Hence, we ask the question whether popular CNNs, when trained for valence prediction, show a similar internal structure. Following this, we turn our attention to the question of what object classes at different network depths have the most influence on the alignment between the CNNs and the human brain, and whether different network architectures show similar behavior in this regard. To answer this question, we introduce Object2Brain, a novel framework to weight the importance of different object classes for the alignment between the CNNs and the human brain. To the best of our knowledge, we are the first to examine CNN-to-human correspondence for the specific case of image valence appraisal.

Concretely, we address the following research questions:
\begin{enumerate}[label={Q\arabic*.}, leftmargin=2em]
\item To what extent does a CNN trained to predict image valence align with human behavioral and fMRI data? Does the layered structure of the CNN mimic the low-mid-high level processing found in the brain?
\item Is this alignment influenced by specific object categories at specific depths of the network? Is behavior consistent across architectures?
\end{enumerate}

The remainder of this paper is organized as follows. \cref{s:corr_to_brain} focuses on Q1. We summarize the work from psychology that forms the basis for our current work, and explore the correlations between CNNs and human data for valence appraisal. Next, we investigate Q2 in \cref{s:inf_obj_classes}, starting with an introduction to the tools used and an in-depth explanation of how we determine the effect of different object categories on the correlations through Object2Brain, followed by a discussion of experimental results. In \cref{s:future_work} we examine weaknesses of our current approach and potential roads for tackling those. Finally, we summarize our findings and conclude in \cref{s:ch4_conclusion}. Space constraints limit us to a few key figures. Our code repository\footnote{\url{https://gitlab.com/EAVISE/lme/object2brain}} contains further plots for all considered architectures and an extended range of parameters.

\section{CNN-to-Human Alignment for Image Valence Appraisal}
\label{s:corr_to_brain}
In this section, we explore the correspondence between different CNNs and the human brain for the task of valence appraisal. We start by summarizing the work in psychology, and the accompanying dataset, on which our work builds. Next, we describe how we checked the alignment of CNNs with the psychology findings, and report our obtained results.

\subsection{fMRI Study}
\label{ss:fmri_data}
Our work builds on results presented in \cite{Yargholi2025}. This work studies the processing of the valence of social scenes by the human brain. A carefully curated dataset of 48 images is introduced, each depicting multiple people in various settings, with balanced conditions of conformity between the people and scene valences. We refer to this dataset as \emph{\fmriimgs{}}. Concretely, the dataset balances between 4 different conditions---12 images for each condition---related to the people in the image (facial expression, body pose; $P$) and the scene elements (anything but the people, including clothing; $S$): \emph{P+S+}, \emph{P+S-}, \emph{P-S-} and \emph{P-S+}, with \emph{+} and \emph{-} indicating positive and negative valence respectively. E.g., a \emph{P+S+} image might represent happy participants at a wedding, while a \emph{P+S-} image might represent playful children in a war ravaged zone. If \emph{P} and \emph{S} agree, the images are referred to as \emph{congruent} (24 images); if they disagree the images are \emph{incongruent} (24 images).

In a first phase, these 48 images were annotated each by 50 annotators for three valence types, using an integer scale ranging from 1 to 7: \emph{image valence} (whole image; IV), \emph{people valence} (PV) and \emph{scene valence} (SV). These annotations constitute the \emph{behavioral} data. For congruent images, PV and SV were similar, while being markedly different for incongruent images, consistent with the dataset setup. From the annotations, binary labels were obtained for each valence type by assigning ``negative'' ($0$) to images whose valence was lower than the median for that type, and ``positive'' ($+1$) to those with valence higher than the median. We refer to these labels as the \emph{true labels}.

In a second phase, using the same images, 22 subjects were placed inside an fMRI scanner and instructed to rate the valence of the entire image on a scale from 1 to 4. The Human Connectome Project brain regions were used for subsequent data analysis. Relevant to this work, these regions were clustered into three large regions, namely, quoting the original text: ``(i) low-level visual regions (retinotopic areas V1-V4), (ii) mid-level association areas mostly in posterior cortex, and (iii) high-level [Regions Of Interest] involved in control processes and mostly in anterior cortex.'' We refer to these as the \emph{LLR}, \emph{MLR} and \emph{HLR} respectively.

To determine what type of valence each of the three aggregate regions is sensitive to, a Multivariate Pattern Analysis approach \cite{2012_fMRIDecoding} was used whereby linear models were trained using ridge regression and using the (average) activations of each large region's subregions as features, and the true IV, PV and SV labels as targets. It was shown that there was a small but significant correlation between predicted and true labels for all valence types in the MLR and HLR, but only for scene valence in the LLR. This suggests that the lower level brain regions are only sensitive to the valence of scene elements, and that higher-order processing is needed to properly assess the people and image valence. Looking deeper at the division between congruent and incongruent images, it was shown that positive correlations between predicted and true labels existed for all valence types and all large regions for congruent images. In contrast, for incongruent images no correlations were observed for any valence type for the LLR, while positive correlations were observed for all valence types and both the MLR and HLR, albeit smaller than for congruent images.

\subsection{Correlation Between CNNs and Humans}
\label{ss:corr_with_fmri}
To assess the alignment between CNNs and humans, we compare the Spearman R correlation between the predictions of several CNN regression models and both the behavioral data and the predictions of the linear models trained on the LLR, MLR and HLR fMRI data introduced in \cref{ss:fmri_data}. In total, this amounts to 24 \emph{correlation targets}: true labels and three model predictions each for IV, PV and SV, further split over the congruent and incongruent images. In a first experiment, we analyze the alignment between several CNN architectures and human data. Next, we limit ourselves to a few architectures, and check whether the way correlations between the layers in a CNN and the human data evolve from early to later layers follows a similar pattern as observed in the fMRI data by examining CNNs trained using features obtained at multiple cut-off points in the network architecture.

\subsubsection{Experimental Approach}
To assess the alignment between a particular trained CNN and the brain, we let it process all \fmriimgs{}, and compute Spearman's R between its predictions and the 24 correlation targets. All models considered in this paper were trained using the FindingEmo dataset \cite{FindingEmo24}, which contains 25,869 images, each depicting multiple people in various settings similar to \fmriimgs{}. Each image is annotated along several dimensions, among which (image) valence ratings on an integer scale from -3 to +3. To have the CNN's target align as closely as possible with the linear models, we assigned a label of 0 to FindingEmo images with negative valence rating (8469 images), 1 to images with a positive valence rating (13997 images), and discarded images with a valence rating of 0 (3403 images). For all training runs, the data was randomly split into a stratified 80/20 train/test split. 

\subsubsection{Correlation With Popular Architectures}
\label{sss:corr_many_archs}
As a first indication of how well CNNs align with the brain for valence prediction, we consider a large set of architectures and compute the correlation between their predictions and the human data. Our analysis includes AlexNet \cite{AlexNet}, DenseNet121 and 161 \cite{DenseNet}, GoogLeNet \cite{GoogLeNet}, ResNet18, 34, 50 and 101 \cite{ResNet} and VGG16 and 19 \cite{VGG}. For all architectures, we use the default ImageNet-1k \cite{ILSVRC15} pretrained PyTorch implementations. Besides these models, we also use the official AlexNet, DenseNet161 and ResNet18 and 50 models trained on the Places365 dataset \cite{zhou2017places}.

For each architecture, we take the pretrained model and replace the last layer with a single output node. Except for this last layer, the model is frozen. Models were trained for a maximum of 250 epochs with a batchsize of $50$, using MSELoss. We used starting learning rate (LR) $\in[0.001, 0.002, 0.0001, 0.0005]$, with LR update rule $\mbox{lr}_{e} = \nicefrac{\mbox{lr}_0}{\sqrt{(e//5) + 1}}$, with $\mbox{lr}_0$ the starting LR, and $\mbox{lr}_e$ the learning rate at epoch $e$. Training stopped when either the test loss had not improved for 8 epochs in a row or the maximum number of epochs was reached, with the best model at that point put forward as the final trained model. For each LR, 5 models were trained. The model achieving the best Mean Average Error (MAE) score over all LRs and runs was kept, resulting in one model per architecture.

\begin{figure*}[!htb]
\centering
\includegraphics[width=\linewidth]{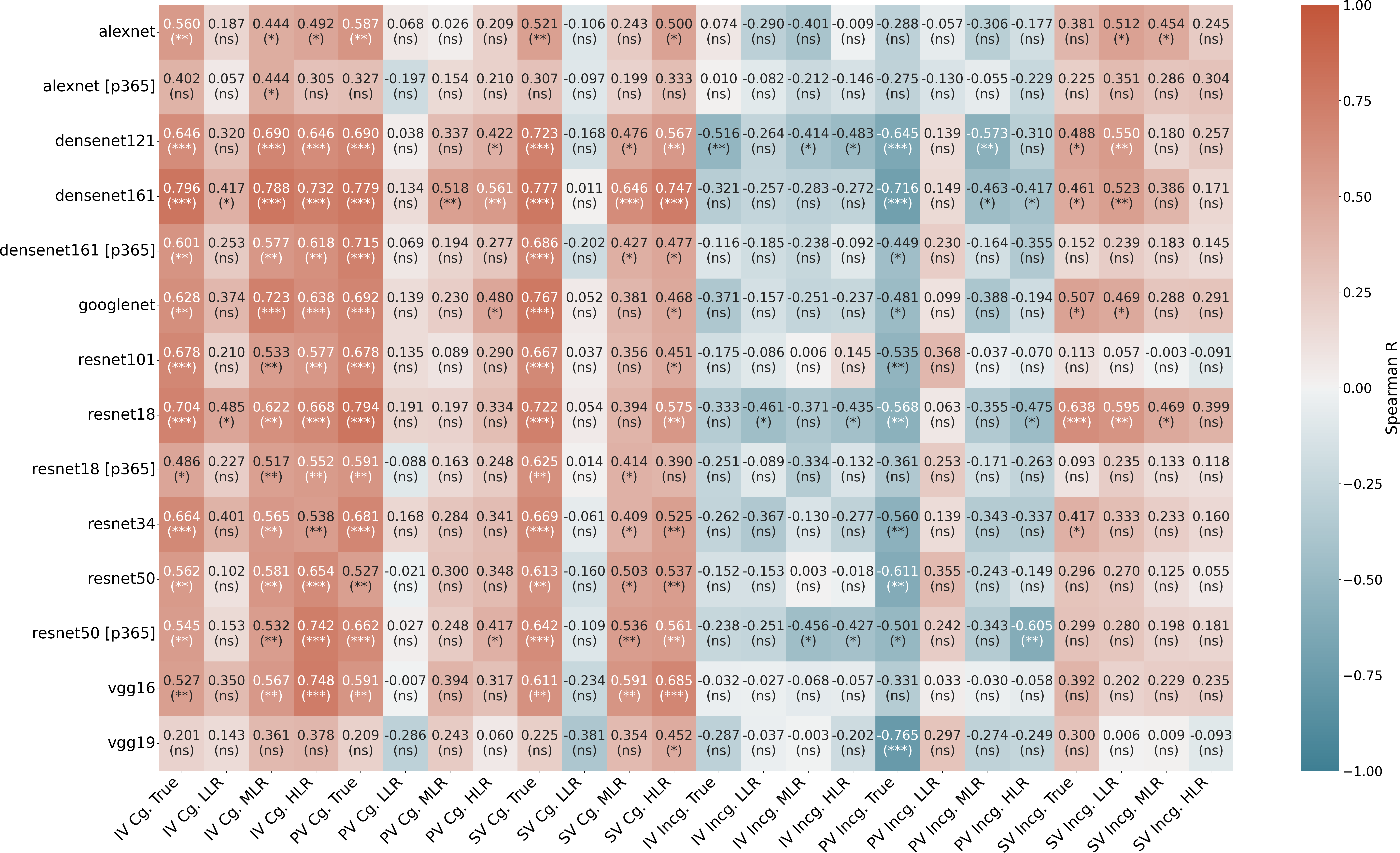}
\caption{Heat map showing the correlations between different CNN architectures, and both behavioral data (``True'') and the output of linear models trained on fMRI data (``LLR/MLR/HLR''). All models based on an ImageNet-trained backbone, except ``[p365]'' models which use a Places365-trained backbone. Significance indicated by: $* = (p<0.05), ** = (p<0.01), *** = (p<0.001)$.}
\label{fig:corr_model_brain_full_models}
\end{figure*}

The correlations between the models' valence predictions on \fmriimgs{} and the human data are grouped in \figref{fig:corr_model_brain_full_models}. The graph is ordered such that all congruent targets occupy the left half, and conversely for the incongruent targets. What is immediately clear is that correlations are significantly higher with congruent targets than with incongruent targets. Correlations tend to be negative with incongruent targets, except for SV targets, indicating an overall strong sensitivity to scene elements of these models. The Places365 backbone does however not translate to higher SV correlations, despite Places365 being a dataset specifically aimed at scene recognition. In effect, the Places365-based models achieve markedly lower correlations overall compared to their ImageNet-based counterparts. Striking are also the (often significant) negative correlations with incongruent PV true labels. Together with the positive correlations for incongruent SV true labels, this strongly suggests the models are not capable of disentangling both elements for valence appraisal, and again suggests the SV is decisive for the model's decision, similar to the LLR.

Deeper architectures do not necessarily perform better than the shallower siblings. DenseNet161 does achieve higher correlations than the 121-layer deep variant for congruent targets, but for both VGG and ResNet the shallowest variants (VGG16 and ResNet18 respectively) achieve the highest correlations.

\subsubsection{Evolution Across Layers}
\label{sss:evo_layers}
For the more in-depth analysis, tracking model alignment across architecture depth, we limit ourselves to only 3 architectures: AlexNet, VGG16 and ResNet18. Although considered older by today's standards, these architectures have been shown to exhibit strong structural hierarchical similarities to the human brain in terms of object representations \cite{Nonaka2020BrainHS}. Moreover, we have shown with \figref{fig:corr_model_brain_full_models} that VGG16 and ResNet18 actually performed best in their respective model families.

\begin{figure*}[htb]
\centering
\includegraphics[width=\linewidth]{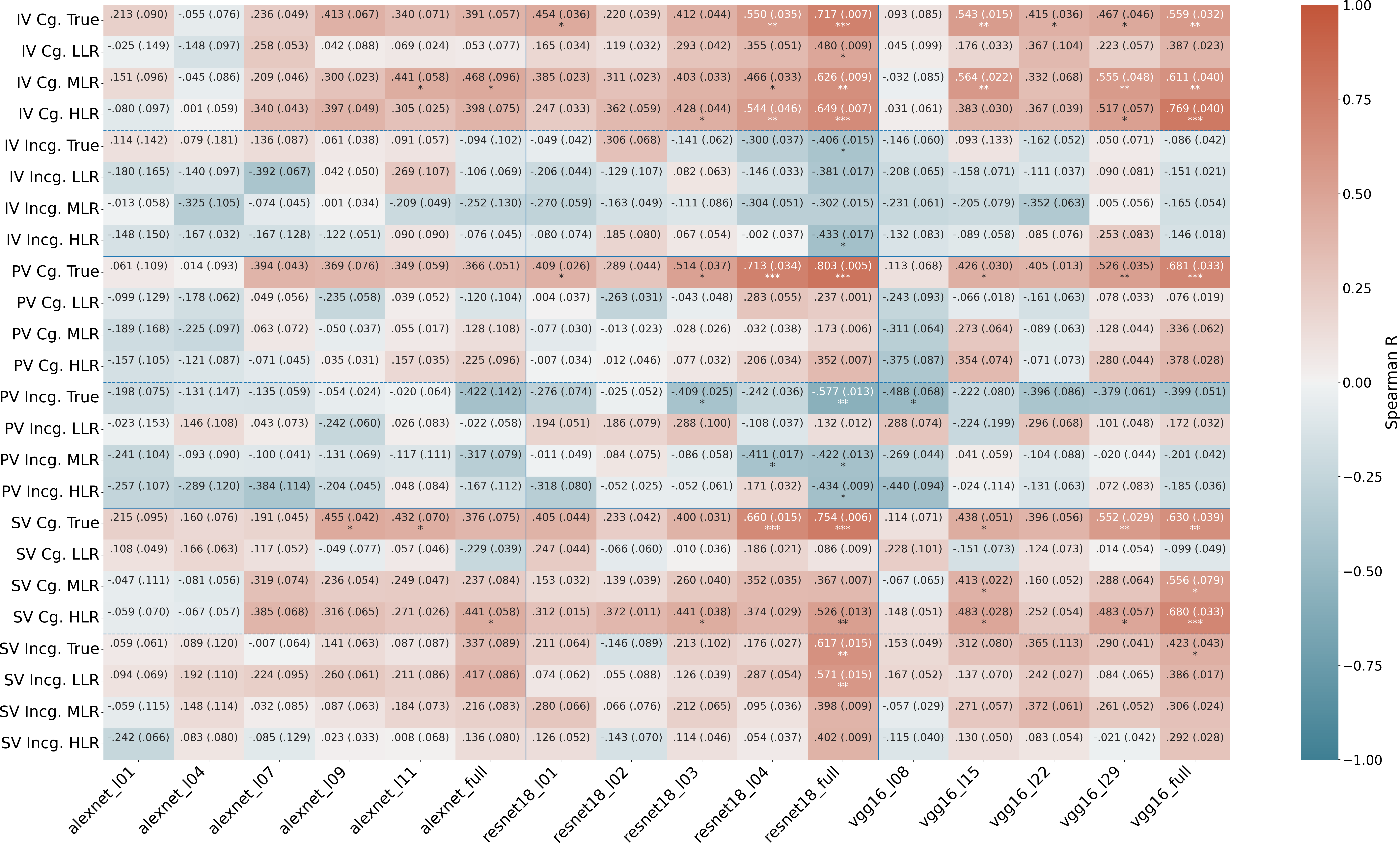}
\caption{Heat map showing the correlations between different ReLU'd convolutional layers and full models in several popular CNNs, and both behavioral data (``True'') and the output of linear models trained on fMRI data (``LLR/MLR/HLR''). Averages shown over 5 runs, with standard deviations between parentheses. For ResNet18, the layer indices refer to groups. Significance indicated by: $* = (p<0.05), ** = (p<0.01), *** = (p<0.001)$.}
\label{fig:corr_model_brain}
\end{figure*}

We again used the default PyTorch implementations of these architectures, pre-trained on ImageNet-1k.  We cut off the networks after each ReLU'd convolutional layer\footnote{I.e., after each rectified linear unit (ReLU) following a convolutional layer.} for AlexNet (layers 1, 4, 7, 9 and 11), ReLU'd convolutional layers preceding a MaxPool operation for VGG16 (layers 8, 15, 22 and 29) and layers 1, 2, 3 and 4 for ResNet18\footnote{For ResNet, these ``layers'' represent groups of two ``BasicBlocks''. They are named ``layer1'' till ``layer4'' in PyTorch, and we follow this naming convention.}, freeze the resulting network and feed the flattened features obtained at that point to a linear layer with a single output node and sigmoid activation function. Only this last linear layer is trained. We also train models using the frozen full architectures, only replacing the last linear layer with a to-be-trained single output node layer. Models were trained for a maximum of 75 epochs with a batchsize of $10$, using WeightedMSELoss \cite{FindingEmo24}. Starting LR were $[2e-6, 1e-6]$ for AlexNet and ResNet18, and $[2e-7, 1e-7]$ for VGG16, with the LR update rule reading $\mbox{lr}_{e} = \nicefrac{\mbox{lr}_0}{\sqrt{(e//3) + 1}}$, with $\mbox{lr}_0$ the starting LR and $\mbox{lr}_e$ the learning rate at epoch $e$.\footnote{For all full models we used a wider range of LR $\in [2e-4, 1e-4, 2e-5, 1e-5, 2e-6, 1e-6]$.} Training stopped when either the test loss had not improved for 6 epochs in a row or the maximum number of epochs was reached, with the corresponding best model put forward as the final trained model. We trained a model for each starting LR, and kept the one with the best MAE score.\footnote{Model performance on the training and test sets are available from our repository.} For each cut-off point, we repeat the experiment 5 times using random seeds, and report averages and standard deviations of the computed Spearman R values. The results are shown in \figref{fig:corr_model_brain}.

Observe that 42 out of the 53 significant correlations are obtained for congruent images, with significant correlations typically being obtained for later layers. For incongruent images, correlations are clearly highest for SV, most often showing (in some cases significant) negative correlations with PV and IV. ResNet18 and VGG16 full models show strong positive correlations with the IV, PV and SV true labels for congruent images, but only with SV true labels for incongruent images, identical to the LLR. For congruent images, within a same valence type correlations tend to increase from LLR to HLR, while for incongruent images the inverse trend is apparent (highest correlation with LLR), particularly for PV, further cementing the likeness between the CNNs and the LLR. Despite the FindingEmo annotations being for IV, the trained models appear to focus on scene elements, again evidenced by the high correlations for incongruent SV.

\section{Influence of Object Classes}
\label{s:inf_obj_classes}
In this section, we explore the effect of several object classes on the correlations of the CNNs with the human data. We start by introducing \emocam{}, i.e., the approach used to relate object classes to CNN filters. We then describe how we combine this information with the correlation analysis in \cref{s:corr_to_brain} to determine the effect of specific classes on specific correlations, and discuss our findings. 

\subsection{\emocam}
\label{ss:emocam}
EmoCAM was introduced in \cite{EmoCAM24} as a pipeline that allows to determine, on an image corpus level, what object classes are most salient to a CNN for predicting a particular output label. In short, EmoCAM combines an object detection network, in casu YOLOv3 \cite{YOLOv3} trained on Open Images v1, with Class Activation Mapping \cite{CAM}, to weight the importance of each object class to a network $N$'s predictions by processing an image corpus $D$ as follows. First, process each image $I \in D$ with YOLO to obtain a list of detected objects and corresponding bounding boxes $B$. Next, let $N$ process $I$, and apply a CAM technique to detect the most salient parts of $I$ to the network's predicted output label, resulting in a heat map $C$. Finally, by comparing the overlap between $C$ and $B$, a score is assigned to each YOLO class respective to $N$'s predicted label, equaling the average pixel activation of the class map within the bounding box of each detected object for that class. Analyzing these scores over the entirety of $D$ allows to determine what object classes were most salient to each output label.

We adapt the EmoCAM approach in several ways to create \emocam{}. First, the system is made to work with Ultralytics\footnote{\url{https://docs.ultralytics.com/}} YOLO models, allowing us to use YOLOv8 trained on Open Images v7.\footnote{At the time of writing, YOLOv8 is the most recent YOLO version for which pretrained weights on Open Images are available.} Second, we update the scoring function to
\begin{equation}
S(b) = \frac{\max(b)}{1 + (\max(b) - \text{avg}(b))},
\end{equation}
with $\max(b)$ the maximum CAM score and $\text{avg}(b)$ the average CAM score within bounding box $b \in B$. This score function gives more weight to bounding boxes with averages close to the maximum, i.e., homogeneous boxes in terms of CAM score. 
 Third, while the original EmoCAM implementation was limited to examining a single convolutional layer, we update the system to be applicable to individual \emph{filters}. Concretely, we apply GradCAM \cite{GradCAM} to a target layer, but do not aggregate the CAM results over all filters, instead further processing each filter individually, allowing to weight the importance of object classes to output labels for each individual filter in the network.\footnote{Although \emocam{} allows the choice of different CAM techniques, it was shown in \cite{EmoCAM24} that results were highly correlated for all choices.} Imagining an emotion detection network, this allows, e.g., to detect filters that are very sensitive to faces when predicting ``Anger''.

\subsection{Object2Brain: Quantizing the Effect of Object Classes on Brain Alignment}
\label{ss:exp_corr_obj_cats}
To explore the effect of specific object classes on the correlations between CNNs and humans, 
 for a given network $N$ and layer $l$, we will first determine the sensitivity of each CNN filter $f^l_i$, with $i$ the index of filter $f$ in layer $l$, to the considered object classes, followed by determining the effect of $f^l_i$ on the correlations with the human data.

For this experiment, we used the ``full'' models from \cref{sss:evo_layers}, repeating the experiment for all 5 models---one per run---per architecture. First, we apply \emocam{} to each network $N$ using the same (ReLU'd) convolutional layers as used in \cref{ss:corr_with_fmri} as target layers, allowing us to score Open Images object classes in relation to each filter in each of these layers. For each targeted layer $l$ of $N$ with $N^l_f$ filters, we thus obtain a matrix $S_l$ of size $N^l_f \times N_C$, with $N_C$ the number of classes. An element $S_l[i,j]$ represents the \emocam{} score of object class $o_j$ for filter $f^l_i$. E.g., $S_4[16,90]$ would represent the score of the 90\supscr{th} class (``Car'') for filter $f^4_{16}$, i.e., the 16\supscr{th} filter in the 4\supscr{th} layer of $N$.

\begin{figure*}[!tb]
\centering
\includegraphics[width=\linewidth]{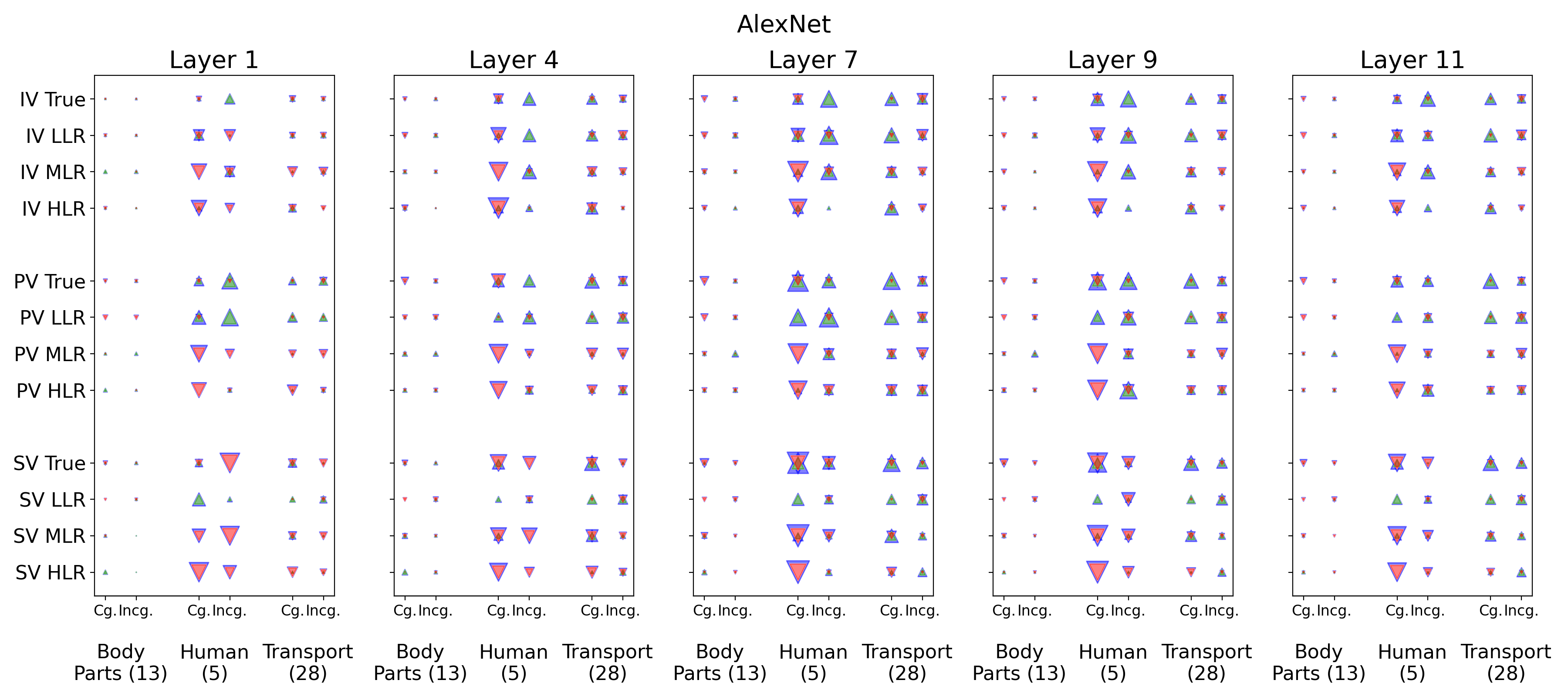}
\caption{Evolution of the average effect of disabling filters sensitive to the ``Body Parts'', ``Human'' and ``Transport'' categories on the correlations between AlexNet and humans across layers; results shown are averages over 5 models. `Cg.' = congruent images, `Incg.' = incongruent images, upward green triangles = positive effect, downward red triangles = negative effect, blue halo = standard deviation. Number of constituent classes in parentheses. Triangle size follows effect size.}
\label{fig:exp2_per_layer}
\end{figure*}

To determine the effect of each filter $f^l_i$ on the correlations, we simply eliminate it (effectively by putting it to an all-zero matrix), and denote the ensuing model as $N\!\setminus\! f^l_i$. Denoting by $t_j$ the $N_t=24$ correlation targets (see \cref{ss:corr_with_fmri}; e.g., ``IV~Cg.~True''), we compute the correlations $c^{t_j}_N$ with target $t_j$ for $N$, and then recompute the correlations for $N\setminus f^l_i$, obtaining $c^{t_j}_{N\setminus f^l_i}$. Repeating this process for each filter, we obtain matrix $C_l$ of size $N^l_f \times N_t$, with entries 
\begin{equation}
C_l[i,j] = c^{t_j}_{f^l_i} - c^{t_j}_{N\setminus f^l_i}.
\end{equation}
Thus, $C_4[0,16]$ contains the difference in correlation with target $0$ (say, ``IV~Cg.~True'') between the base network $N$ and the same network with the 16\supscr{th} filter in the 4\supscr{th} layer replaced by an all-zero matrix.

Now we combine both pieces of information---$S_l$ and $C_l$---to obtain weights representing how much each \emph{object class} influences the correlations. To this end, we start by multiplying $S_l$ with $C_l$ to obtain a matrix $W_l$ of size $N_t \times N^l_f \times N_C$ as follows:
\begin{equation}
W_l[i,j,k] = C_l[j,i] \times S_l[j,k].
\end{equation}
In other words, $W_l[i,j,k]$ represents the effect of eliminating filter $f^l_j$ on the correlations with target $t_i$ weighted by the sensitivity of filter $f^l_j$ to object class $o_k$. Each submatrix $W^{t_i}_l = W_l[i={t_i},j,k]$ thus represents, for layer $l$ and correlation target $t_i$, a matrix that assigns a score to each object class $o_k$ and filter $f^l_j$. E.g., $W_4^0[16,90] = W_4[0,16,90]$ encodes the effect of class $o_{90}=\mbox{``Car''}$ in the 16\supscr{th} filter of layer 4 to the correlation with $t_0=\mbox{``IV~Cg.~True''}$.

Finally, to determine the total effect of object class $o_k$ in layer $l$ on the correlation with target $t_i$, we sum $W^{t_i}_l$ over all filters and obtain vector $V^{t_i}_l$ of size $N_C$, whose elements, defined as
\begin{equation}
V^{t_i}_l[k] = \sum_{j=1}^{N_f} W^{t_i}_l[j,k],
\end{equation}
represent weights indicating \emph{the importance of object class $o_k$ on the correlation with target $t_i$} taken over all filters in layer $l$. Keeping with our running example, $V^{t_0}_4[90]$ represents the $overall$ importance of object class $o_{90}=\mbox{``Car''}$ in layer 4 on the correlation with target $t_0=\mbox{``IV~Cg.~True''}$.

We now apply Object2Brain using the valence prediction CNNs trained on FindingEmo and the correlations with the human behavioral and fMRI data. I.e., we use FindingEmo to obtain $S_l$ (with only one output label, i.e., valence), and use the human data to obtain $C_l$.

To facilitate interpretation of the results, we grouped the 601 Open Images classes into 34 categories. The full mapping is available from our repository; we limit ourselves here to the categories most relevant to the analysis, which are (number of constituent classes and examples in parentheses): ``Clothing'' (30; ``Backpack'', ``Dress'', ``Jeans''), ``Furniture'' (25; ``Couch'', ``Lamp'', ``Table''), ``Health'' (9; ``Band-aid'', ``Crutch'', ``Toothbrush''), ``Body Parts'' (13; ``Human arm'', ``Human ear'', ``Human head''), ``Human'' (5; ``Boy'', ``Girl'', ``Man'', ``Woman'' and ``Person''), ``Nature'' (14; ``Honeycomb'', ``Maple'', ``Rose''), ``Places'' (11, ``Fountain'', ``Lighthouse'', ``Skyscraper''), ``Sports'' (39; ``Canoe'', ``Jet ski'', ``Racket'') and ``Transport'' (28; ``Bus'', ``Helicopter'', ``Segway''). To analyze the data, we sort $V^{t_i}_l$, essentially ordering the classes according to weight per correlation target. We then take the top $X$ highest and lowest ranked classes, and sum their weights over their respective categories to obtain the categories that contribute most positively and negatively respectively. We are interested in the relative sum values between categories. For unconsidered categories, the contributions were most often close to $0$. This is likely caused by the fact that only 250 out of the 601 Open Images classes were detected in the FindingEmo dataset.

\begin{figure*}[!p]
\centering
\includegraphics[width=\linewidth]{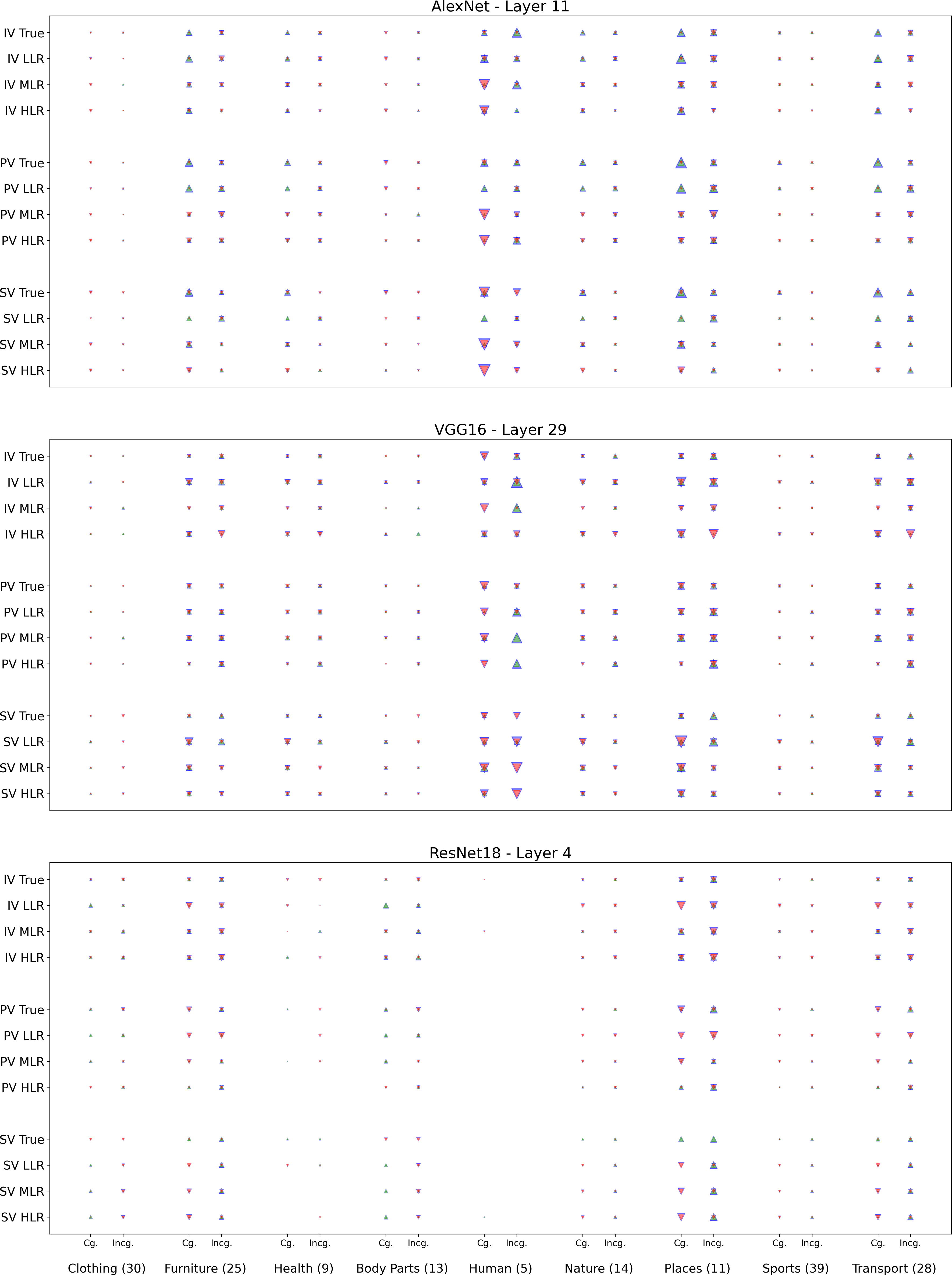}
\caption{Average effect of disabling category sensitive filters on the correlations between the last cut-off layer of different CNN architectures and humans; results shown are averages over 5 models. `Cg.' = congruent images, `Incg.' = incongruent images, upward green triangles = positive effect, downward red triangles = negative effect, blue halo = standard deviation. Number of constituent classes in parentheses. Triangle size follows effect size.}
\label{fig:exp2_per_arch}
\end{figure*}

Using $X=25$ (i.e., 10\% of the detected classes) and further limiting ourselves to three categories (``Body Parts'', ``Human'' and ``Transport''), \figref{fig:exp2_per_layer} shows the evolution of the category weights, averaged over their constituent classes, across layers in AlexNet.\footnote{Refer to our repository for analogous plots for different values of $X$, as well as plots per category, also for VGG16 and ResNet18.} A same category can have both a positive and a negative contribution by virtue of some of their constituent classes contributing positively, others negatively. As can be seen, the ``Body Parts'' and ``Humans'' categories often mirror each other for PV: if one contributes positively, the other tends to contribute negatively. Given that ``Body Parts'' is physically a subset of ``Humans'', this is quite unexpected. From layer 7 onward AlexNet shows close to a steady state at the category level, with even layers 1 and 4 showing clear similarities. This suggests that it does not really matter where exactly in the model sensitivity to a certain category is reduced by eliminating corresponding filters. The overall low variability might be explained by all models sharing the same pretrained backbone. Some evolutions are however noteworthy. E.g., ``PV LLR/Human/Incg.'' starts out strongly positive in layer 1, but ultimately ends up almost equally positive as negative in layer 11.

Not shown here, VGG16 exhibits remarkable consistency across all 4 considered target layers, even though effect sizes differ significantly between layers, clearly peaking in layer 22. For ResNet18 the story is slightly different. Strikingly, this architecture shows negligible sensitivity to the ``Humans'' category. Also, the effect size is generally markedly lower than for the other two architectures. Overall however, just like for AlexNet and VGG16, behavior between layers shows strong consistency averaged over categories.

Turning to comparing the architectures, \figref{fig:exp2_per_arch} groups the averaged results for $X=25$ and the 9 most salient categories for the last cut-off layers of AlexNet, VGG16 and ResNet18.\footnote{For other layers and plots per category, see our repository.} In general, the size of the effect does not necessarily correlate with the correlations show in \figref{fig:corr_model_brain}. Clear distinctions do however manifest themselves between valence types. E.g., the ``Human'' category clearly affects correlations differently for incongruent IV and PV on the one hand and SV on the other, for both AlexNet and VGG16.

\begin{figure}[!htb]
\centering
\includegraphics[width=0.6\columnwidth]{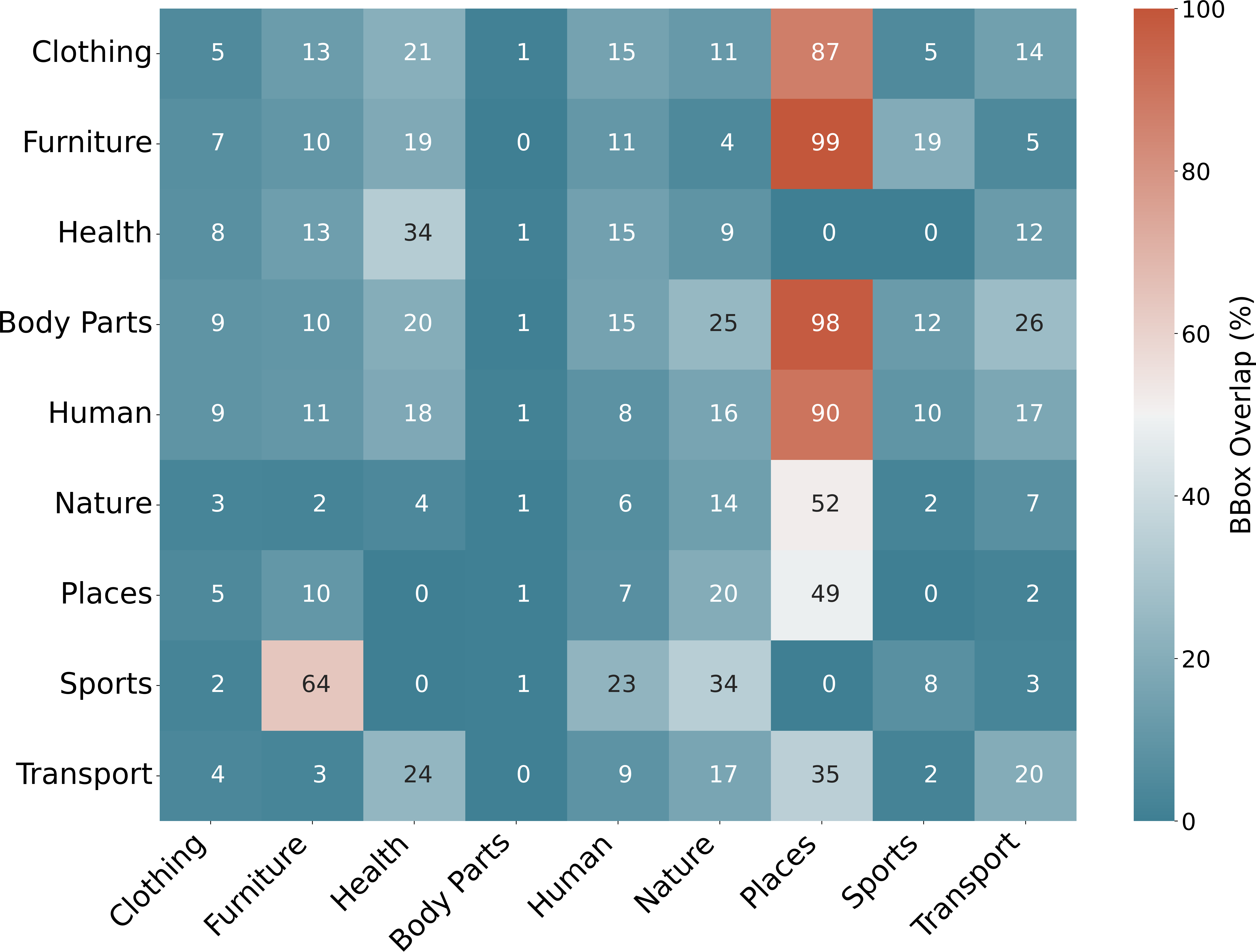}
\caption{Average overlap between bounding boxes of different categories in the FindingEmo dataset. E.g., read entry (row=``Human'', col=``Places'') as: when objects of categories ``Human'' and ``Places'' appear together, on average the intersection of their bounding boxes equals 90\% of the area of the ``Human'' object.}
\label{fig:heatmap_bb_overlap}
\end{figure}

\figref{fig:exp2_per_arch} does make clear that all networks show idiosyncratic sensitivities to the explored categories, even though clear tendencies appear in terms of effect magnitude. Indeed, ``Furniture'', ``Places'' and ``Transport'' show clearly larger effect sizes for all three architectures, and the ``Human'' category is very clearly the most sensitive category for both AlexNet and VGG16. We investigated whether this might be a result of strongly overlapping bounding boxes between classes. E.g., people in cars might result in a large overlap between categories ``Transport'' and ``Human'', thus explaining the large effect for the ``Transport'' category. \figref{fig:heatmap_bb_overlap}, which shows the average bounding box area overlap between any two categories in the FindingEmo dataset, demonstrates this explanation is insufficient. It indeed shows a strong overlap between ``Furniture'' and ``Places'' (99\%), but much smaller overlap between ``Transport'' and ``Places'' (35\%). Moreover, ``Body Parts'' overlaps 98\% with ``Places'', but shows no effect correlation with, e.g., ``Furniture''.

Overall, through Object2Brain it is clear that different architectures appear to process information differently. They exhibit different object class sensitivities, despite showing strong similarities in terms of correlation patterns with human behavioral and fMRI data. For all architectures, the bias toward SV described in \cref{ss:corr_with_fmri} is evidenced by the strong sensitivity to contextual elements. Our method also clearly shows how correspondence with congruent and incongruent images is affected differently.

\section{Future work}
\label{s:future_work}
Our current implementation demonstrates the feasibility of our approach, but presents several opportunities for improvement. For one thing, our approach can only be as good as \emocam{} is precise. Currently however, \emocam{} considers all detected objects whose confidence exceeds a given threshold (we use the default YOLOv8 value of 0.25), but does not take this confidence into account to compute object scores. Doing so might prove worthwhile.

The biggest drawback of our approach is the use of bounding boxes. Although we have argued in \cref{ss:exp_corr_obj_cats} that a straightforward relationship between class overlap and correlations with human data is absent, contamination can not be excluded. A solution would be to use a segmentation approach, but then the limited classes of the most popular segmentation models become an issue. Still, as a first step in this direction, a combination of both an object detection network (e.g., YOLOv8 + Open Images) and a segmentation network (e.g., YOLOv11 + COCO) could be envisaged.

\section{Conclusion}
\label{s:ch4_conclusion}
We have investigated the alignment between multiple CNN architectures, trained for the task of valence appraisal from static images, to human behavioral and fMRI data. Our results suggest that, for this particular task, these CNNs show most commonality with the low-level visual brain regions, evidenced by a shared bias toward scene valence. This suggests that more complex network architectures might be needed to faithfully simulate human valence appraisal, specifically to go beyond the higher cognitive brain processing.
Furthermore, we presented Object2Brain, a novel framework that combines GradCAM and object detection at the CNN-filter level to study the influence of different object classes on CNN-to-human alignment. We have shown that despite overall similarities in correlation trends, different architectures appear to process information in distinct ways with distinct object class sensitivities. This opens the door to exciting new analyses of which CNNs best correspond to human behavior, and how existing architectures can be adapted to improve alignment.

\begin{ack}
This work was funded by KU Leuven grant IDN/21/010.
\end{ack}

\bibliographystyle{ieeetran}
\bibliography{object2brain}

\end{document}